\documentclass[10pt,twocolumn,letterpaper]{article}

\usepackage{cvpr}
\usepackage{subfig}
\usepackage{times}
\usepackage{xspace}
\usepackage{times}
\usepackage{epsfig}
\usepackage{graphicx}
\usepackage{amsmath}
\usepackage{amssymb}
\usepackage{algorithm}
\usepackage{algorithmic}
\usepackage{mathtools}
\usepackage{array}
\usepackage{multirow}
\usepackage[inline]{enumitem}
\usepackage{subfig}
\usepackage{multirow}
\usepackage{color}
\usepackage[utf8]{inputenc} 
\usepackage[T1]{fontenc}    
\usepackage{url}            
\usepackage{booktabs}       
\usepackage{amsfonts}       
\usepackage{nicefrac}       
\usepackage{microtype}      
\usepackage{xcolor}
\newcommand{\softmax}{\text{softmax}}

\usepackage[breaklinks=true,bookmarks=false]{hyperref}

 \cvprfinalcopy 

\ifcvprfinal\fi
\begin{document}

\title{Understanding Visual Ads by Aligning Symbols and Objects using Co-Attention}

\author{
	Karuna Ahuja \quad Karan Sikka \quad Anirban Roy \quad Ajay Divakaran \\ {\tt\small (karuna.ahuja, karan.sikka,
	anirban.roy, ajay.divakaran)@sri.com} \\ SRI International, Princeton, NJ
}

\maketitle

\makeatletter
\DeclareRobustCommand\onedot{\futurelet\@let@token\@onedot}
\def\@onedot{\ifx\@let@token.\else.\null\fi\xspace}

\def\etal{et al\onedot}
\def\etc{etc\onedot}
\def\ie{i.e\onedot}
\def\eg{e.g\onedot}
\def\cf{cf\onedot}
\def\vs{vs\onedot}
\def\pd{\partial}
\def\grad{\nabla}
\def\Li{\mathcal{L}}
\def\O{\mathcal{O}}
\def\N{\mathbb{N}}
\def\C{\mathcal{C}}
\def\Lt{\tilde{L}}
\def\R{\mathbb{R}}
\def\X{\mathcal{X}}
\def\I{\mathcal{I}}
\def\F{\mathcal{F}}
\def\w{\textbf{w}}
\def\x{\textbf{x}}
\def\k{\textbf{k}}
\def\k{\textbf{k}}
\def\d{\boldsymbol{\delta}}
\def\y{\textbf{y}}
\def\l{\boldsymbol{\ell}}
\def\wrt{w.r.t\onedot}
\def\a{\boldsymbol{\alpha}}
\def\vertspace{0.6em}

\definecolor{redcol}{rgb}{1, 0, 0}
\definecolor{bluecol}{rgb}{0, 0, 1}
\newcommand{\red}[1]{\textcolor{redcol}{#1}} 
\newcommand{\blue}[1]{\textcolor{bluecol}{#1}} 

\begin{abstract}
We tackle the problem of understanding visual ads where given an ad image, our goal is to rank appropriate human
	generated statements describing the purpose of the ad. 
	This problem is generally addressed by jointly embedding images  and candidate statements to establish
	correspondence.  Decoding a visual ad requires inference of both semantic and symbolic nuances referenced in an image
	and prior methods may fail to capture such associations especially with weakly annotated symbols.  In order to
	create better embeddings, we leverage an attention mechanism to associate image proposals with symbols and thus
	effectively aggregate information from aligned multimodal representations.  We propose a multihop co-attention
	mechanism that iteratively refines the attention map to ensure accurate attention estimation.  Our attention
	based embedding model is learned end-to-end guided by a max-margin loss function. We show that our model 
	outperforms other baselines on the benchmark Ad dataset and also show qualitative results to highlight the
	advantages of using multihop co-attention.

\end{abstract}


\section{Introduction}

We address the problem of understanding visual advertisement which is a special case of visual content analysis
\cite{hussain2017automatic}. While current vision approaches can successfully address object \cite{krishna2017visual,
kiros2014unifying, lin2014microsoft} and scene \cite{zhou2014learning, girshick2014rich, cordts2016cityscapes} centric
interpretations of an image, deeper subjective interpretations such as rhetoric, symbolism, \etc. remain
challenging and have drawn limited attention from the vision community. 

\begin{figure}[ht]
	\centering
	\includegraphics[width=\columnwidth]{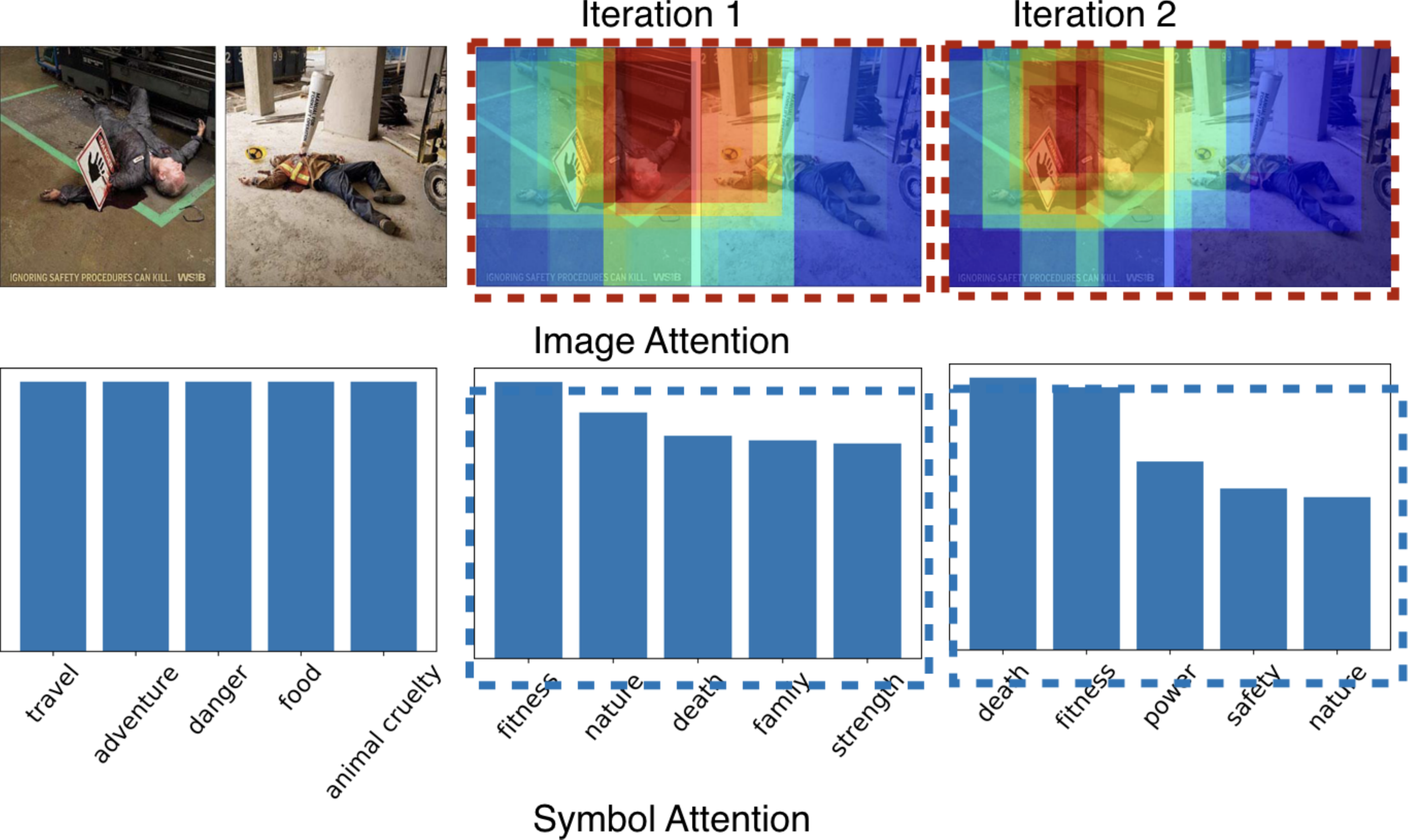} 
\caption{\small{Figure shows the image and symbol attention on a PSA ad about the importance of following safety laws. Our
algorithm learns to iteratively infer the reference of the symbol `death' with the relevant image content.}}
	\label{vis}
\end{figure}

Recently, visual ad-understanding has been addressed by Hussain et al. \cite{hussain2017automatic} and a dataset has
been
introduced to evaluate ad-understanding approaches. The authors introduce several sub-problems such as identifying the
underlying topics in the ad, predicting the sentiments and symbolism referenced in the ads,  associating an action and
corresponding reason in response to an ad.  In this work, we target understanding of ad images by formulating the task
as mutimodal matching of the image and its corresponding human generated statements \cite{ye2017advise}. These
statements were obtained from annotators by asking them to answer questions such as "I should do this because".
\cite{hussain2017automatic}. Note that for interpreting the rhetoric conveyed by an ad image, we need to exploit the
semantic, symbolic, and sentimental references made in the image. Moreover, these alternate references are correlated and
influence each other to convey a specific message. For example, an ad corresponding to symbols `humor' and `fun' are
likely to invoke `amused' sentiment rather than `sad'. Thus, we consider the interactions between these references for
interpreting ads in the proposed approach.

Currently, the amount of labeled data for the task of understanding ads is limited as annotating images with symbols and
sentiments is both subjective and ambiguous.
To tackle these challenges, we propose a novel weakly supervised learning (WSL) algorithm that learns to
effectively combine multiple references present in an image by using an iterative co-attention mechanism. In this work,
we focus only on semantic references (made via visual content) and symbolic references, and later discuss ideas for
including sentiments and object information within our model.

We first obtain scores for symbolic and other references made in an image by using pretrained model trained on
labeled data \cite{hussain2017automatic}. These scores describe symbols at an image level instead at a region-level
granularity due to the difficulty of labeling region-symbol associations. This is often referred to as WSL setting
\cite{roy2017combining, durand2016weldon, kar2017adascan} and
poses specific challenges in understanding ads as different regions are generally associated with different symbols. As
previously mentioned, we pose decoding ads as a multimodal matching problem and use Visual Semantic Embeddings (VSE)
\cite{kiros2014unifying} to jointly embed an image and sentence to a common vector space for establishing correspondence.
However, due to the ambiguity of region-label associations, VSE may fail to correctly align visual regions with symbolic
references and thus unable to fuse information in an optimal manner for decoding the ad.  This motivates us to leverage an
attention driven approach \cite{bahdanau2014neural, xu2015show}, where the prediction task is used to guide the
alignment between the input modalities by predicting attention maps. For example, in
Visual Question Answering (VQA) the task of predicting answers is used to train the question to image attention module
\cite{lu2016hierarchical, anderson2017bottom, das2017human}.

Attention has been shown to improve tasks such as VQA \cite{lu2016hierarchical}, object and scene
understanding \cite{roy2017combining, durand2016weldon}, action recognition \cite{kar2017adascan, sharma2015action,
girdhar2017attentional}. Commonly used top-down attention identifies the discriminative regions of an image based on the
final task \cite{ye2017advise, teh2016attention}.  In ad understanding, image regions may be associated with
multiple symbols. Thus the standard top-down attention may get confused due to the many-to-many mappings between image
regions and image labels. To address this issue, we consider co-attention \cite{lu2016hierarchical, nam2016dual} to
implement an alternating attention from a set of image-level symbols to image regions and vice-versa.  Moreover, recent works demonstrate that the attention maps can be refined in subsequent iterations
\cite{nam2016dual, xu2016ask, yang2016stacked}. Thus,  we consider multi-hop attention (fig.~\ref{vis}) between image regions and symbols where the attention is computed iteratively while attention estimation in current step depends on the
attention from the previous step. We finally fuse information from attended image and symbols from different iterations
for multimodal embedding. We also leverage bottom-up and top-down attention \cite{ye2017advise, anderson2017bottom} by
estimating attention on object proposals leading to improved alignments.  Our work differs from the work by Ye \etal
\cite{ye2017advise} which uses (top-down) attention to attend to image regions and combine additional information
from other references by simple fusion and additional learning constraints. Moreover, our model is principled in using
attention to fuse information from visual and other modalities by using co-attention with multiple hops. Our initial
experiments show that adopting co-attention with multiple hops outperforms the standard top-down and bottom-up
attention in terms of overall ad-understanding performance.


\section{Approach}

\begin{figure}[ht]
	\centering
	\includegraphics[width=\columnwidth]{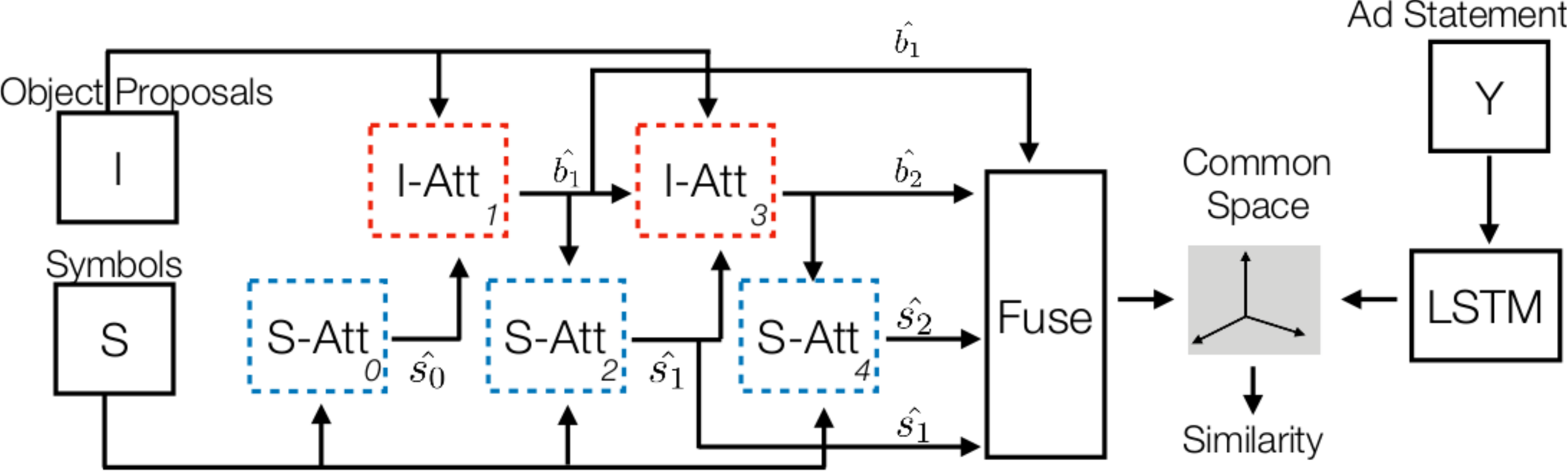} 
	\caption{\small{Figure (best seen in color) shows a block diagram of our algorithm (VSE-CoAtt-2) that uses co-attention with multiple hops
	between visual and symbol modalities. The blocks in red and blue compute attention for image and symbols
	respectively by using the attended vector from other modality.}} 
	\label{block}
\end{figure}

We propose a co-attention based VSE approach, referred to as VSE-CoAtt, for jointly embedding advertisements and their
corresponding ad messages. VSE-CoAtt estimates attention for image regions by using symbol information and subsequently
uses the attended image to predict attention for image symbols. This co-attention formulation allows our method to align
visual and symbol modalities and fuse them effectively. We also propose a multihop version of our algorithm, referred to
as VSE-CoAtt-2, that iteratively refines attention masks for the two modalities (visual and symbolic) and summarizes the
information to compute similarity with an ad statement (fig.~\ref{block}).  We denote an ad image as $I \in
\mathcal{R}^{M \times N \times 3}$. The given ground-truth statements are denoted as $Y=\{y_j\}_{j=1}^{N_m}$.  We use an
embedding to represent each word and use an LSTM \cite{hochreiter1997long} to encode a sentence, denoted as $\psi(y_j)
\in R^{D_3}$. We use object proposals, denoted $\{b_i\}_{i=1}^{N}, b_i \in \mathcal{R}^{4}$ and $N=70$, to attend to
salient image regions 
\cite{anderson2017bottom}.  We use the curated list of $53$ symbols, denoted as $Z=\{z_k\}_{k=1}^{K}, z_k \in
\mathcal{R}^{D_1}$, as provided by the authors of the dataset \cite{hussain2017automatic} and encode them using GloVe vector
\cite{pennington2014glove}. We also assume that we have scores ($p_k$ for symbol $z_k$) either provided by a human
annotator or predicted using another CNN. We use a CNN to extract features from each bounding box $b_i$ and denote them
as $\phi(b_i) \in R^{D_2}$. We begin the iterations for our model by initializing the attended vector (also referred to
as summary vector) for symbols (denoted as $\hat{s_0}$) by:

\begin{align}
\hat{s_0} = \frac{1}{K} \sum_k{z_k}
	\label{eq1}
\end{align}

We compute the raw attention scores for the object proposals $b_i$ by using the attended symbol vector $\hat{s_0}$ as
shown below.  We use $\softmax$ to normalize the attention scores, denoted as $\alpha^{I}_{i}$, and finally compute the
summary vector $\hat{b_1}$ for images

\begin{align}
	\alpha^{I}_{i} &= \softmax(\tanh(\hat{s_0}^TW^T\phi(b_i)) \\
\hat{b_1} &= \sum_i{\alpha^{I}_{i} \phi(b_i)}
\end{align}

where $W \in {D_2 \times D_1}$ is used to project visual features to the symbol space. We use a number subscript in
$\hat{s}$ and $\hat{b}$ to denote the iteration index for the multihop version. We use a similar operation to compute
attention $\beta_{k}^{Z}$ for symbol $z_k$ using the previously attended image vector $\hat{b}_1$ as shown below:

\begin{align}
	\beta^{Z}_{k} &= \softmax(\tanh(z_k^{T} W^T \hat{b_1})) 
\end{align}

We use the given symbol probabilities to weigh the attention maps so as to focus on symbols present in the image as shown below:

\begin{align}	
\hat{s_1} &= \sum_k{\beta^{Z}_{k} p_k z_k}
\end{align}

We consider co-attention with only two hops to avoid overfitting. We obtain the final features for visual and
symbol modalities by fusing the attended vectors at different iterations using an addition operation \ie $f_{IZ} =
\sum_t W^T \hat{b_t} + \hat{s_t}$. Similar to Kiros \etal \cite{kiros2014unifying}, we first linearly project $f_{IZ}$
and then use cosine similarity to compute similarity $S_l$ between $f_{IZ}$ and the $l^{th}$ ad statements $y_{l}$. In
order to learn the model parameters, we use a max-margin based ranking loss which enforces the matching score of an image-symbol
pair to be higher with its true sentences and vice-versa. We define loss for a training sample pair $I, Z$ with
ground-truth ad messages $Y$ as:

\begin{align}
	\mathcal{L}(I, Z, Y, \theta) = \sum_{y_j \in Y} \sum_{y_l \notin Y} \max(0, m - S_{j} + S_{l})
\end{align}

\section{Experiments}

\textbf{Dataset:} Following Ye \etal \cite{ye2017advise}, we evaluate the task of visual ad understanding by
matching ad images to their corresponding human generated sentences on the Ads dataset \cite{hussain2017automatic}.  We
follow the data splits and experimental protocol used by Ye \etal \cite{ye2017advise} and rank 50 statements (3 related
and 47 unrelated from the same topic).  Since the proposed model explores the possibility of including additional
knowledge, we evaluate our approach on a subset of the ADs dataset which have at least one symbol annotation that belongs
to one of the $53$ clusters as in \cite{hussain2017automatic}.
Different from Ye \etal, we make no distinction between the public service announcements (PSAs) and product ads and
combine them during evaluation. During evaluation, we rank the $50$ statements for each image based on their similarity
score and report mean of the top rank of the ground-truth statements for images (mean rank metric). Our dataset consists
of $13,938$ images partitioned into $5$ cross-validation splits- provided by \cite{ye2017advise}.

\textbf{Implementation details:} We extract the features from images (and boxes) using ResNet-101 and
consider top $70$ object proposals \cite{zitnick2014edge}.  We experimented with regions proposals
trained on the symbol bounding boxes, as in \cite{ye2017advise} but found the performance to be lower. For
learning, we use an Adam \cite{kingma2014adam} optimizer with a learning rate of $5e^{-4}$. We implement several
baselines as shown in tab.~\ref{results_1} and use the same features and learning settings for a fair comparison. The
baseline VSE model \cite{kiros2014unifying} with attention (VSE-Att) and without attention (VSE) use features before and after the average pooling layer respectively. Since our model builds on a bottom-up and top-down attention framework, \cite{anderson2017bottom} that uses
object proposals instead of feature maps for attention, we also implement two variants of VSE with object proposals:
1) using average pooling (VSE-P) and 2) using attention over the proposals (VSE-P-Att) (similar to 
Ye \etal \cite{ye2017advise}). We implement four variants of our algorithm that include a single hop co-attention (VSE-CoAtt),
a two hop co-attention (VSE-CoAtt-2), and two similar implementations (VSE-CoAtt-wt and VSE-CoAtt-2-wt) that weigh the symbol initialization (eq.~\ref{eq1}) with symbol probabilities $p_k$. 

\subsection{Results and Ablation Studies}

As show in the Tab.~\ref{results_1}, proposed VSE-CoAtt-2 outperforms all the baseline which justifies the importance of
multihop co-attention for ad-understanding. For example, considering attention cues, VSE-CoAtt-2 achieves a lower mean rank of $6.58$ than VSE (mean rank $7.79$) which does not consider attention. The advantage of using
co-attention is evident in the performance of VSE-CoAtt (rank of $6.68$) versus VSE-P-Att (rank of
$7.35$), that uses a fixed attention template for the visual modality. We also observe benefit of using multiple hops, that
aggregates information from multiple steps of visual and symbol attention, while comparing the mean rank of $6.68$ of
VSE-CoAtt versus mean rank of $6.58$ of VSE-CoAtt-2. The results while using per-symbol probabilities for initializing
iterations for attention seem to be lower for both with and without multihop attention. This could be happening due to
overfitting since we only have a few number of images.

\begin{table}[]
		\centering 
	\resizebox{0.8\linewidth}{!}{
	\begin{tabular}{|l|ccc|c|} 
		\hline 
		METHOD & Box & Att. & Co-& Mean Rank \\ 
		& Proposals &  & Att. &  \\ \hline 
		VSE & & & & $7.79$   \\ \hline 
		VSE-Att & & \checkmark & & $8.32$   \\ \hline 
		VSE-P & \checkmark & & & $7.74$   \\ \hline 
		VSE-P-Att & \checkmark & \checkmark & & $7.35$      \\ \hline \hline 
		VSE-CoAtt & \checkmark & \checkmark & \checkmark  & $6.68$        \\ 
		VSE-CoAtt-2 & \checkmark & \checkmark & \checkmark  & $\textbf{6.58}$        \\ \hline 
		VSE-CoAtt-wt & \checkmark & \checkmark & \checkmark  & $6.77$        \\ 
		VSE-CoAtt-2-wt & \checkmark & \checkmark & \checkmark  & $6.75$        \\ \hline 
	\end{tabular}}
\caption{\small{Comparison of our method (VSE-CoAtt and VSE-CoAtt-2) with different baselines.}}
	\label{results_1}
\end{table}


\section{Conclusion and Future Work}

We propose a novel approach leveraging multihop co-attention for understanding visual ads. Our model uses multiple
iterations of attention to summarize visual and symbolic cues for an ad image. We perform multimodal
embedding of images and statements to establish their correspondence. Our experiments show the advantages of multihop
co-attention over vanilla attention for ad-understanding. 

Beyond the presented work, we are currently working on incorporating additional cues such as sentiments and
objects inside our model. To resolve the problem of limited training data for subjective references, we are using
weakly labeled data on the internet to train models and form associations between objects and symbols. We plan to use
these associations to regularize the predicted attention by our model.

{\small
\bibliographystyle{ieee}
\bibliography{egbib}
}

\newpage
\onecolumn
\section{Appendix}
In this section we show some additional results and provide some more details about the algorithm. We display results for four visual ads corresponding to fast food, road violence, gun violence, and road safety respectively as shown in figure.\ref{fig_all}. We observe that our algorithm is able to both identify as well as refine symbol and image attention in multiple iterations. For \eg, in the advertisement on road safety (last row), the algorithm refines symbol attention in the second hop by shifting attention from unrelated symbols like `power' and 'hunger' to relevant symbols such as `danger' and `safety' over the course of multihop iterations.  

Weakly supervised algorithms often suffer from the problem of identifying the most discriminative region(s) in an image.
Hence, they may fail to cover all possible salient regions and result in overfitting. To prevent this problem, we apply
a heuristic technique wherein for a given $n^{th}$ iteration, we suppress the image attention scores (prior to softmax
operation) of the regions which received scores greater than or equal to $0.7$ times the highest score in ${n-1}^{th}$
iteration. We manually set the scores to a low value ($-2$). Although this simple step improved the results, we need to further investigate the use of additional constraints ( \eg spatial and semantic) to discover other salient regions.


\begin{figure}[ht]
	\centering
	\label{fig_all}
	\includegraphics[width=.5\columnwidth]{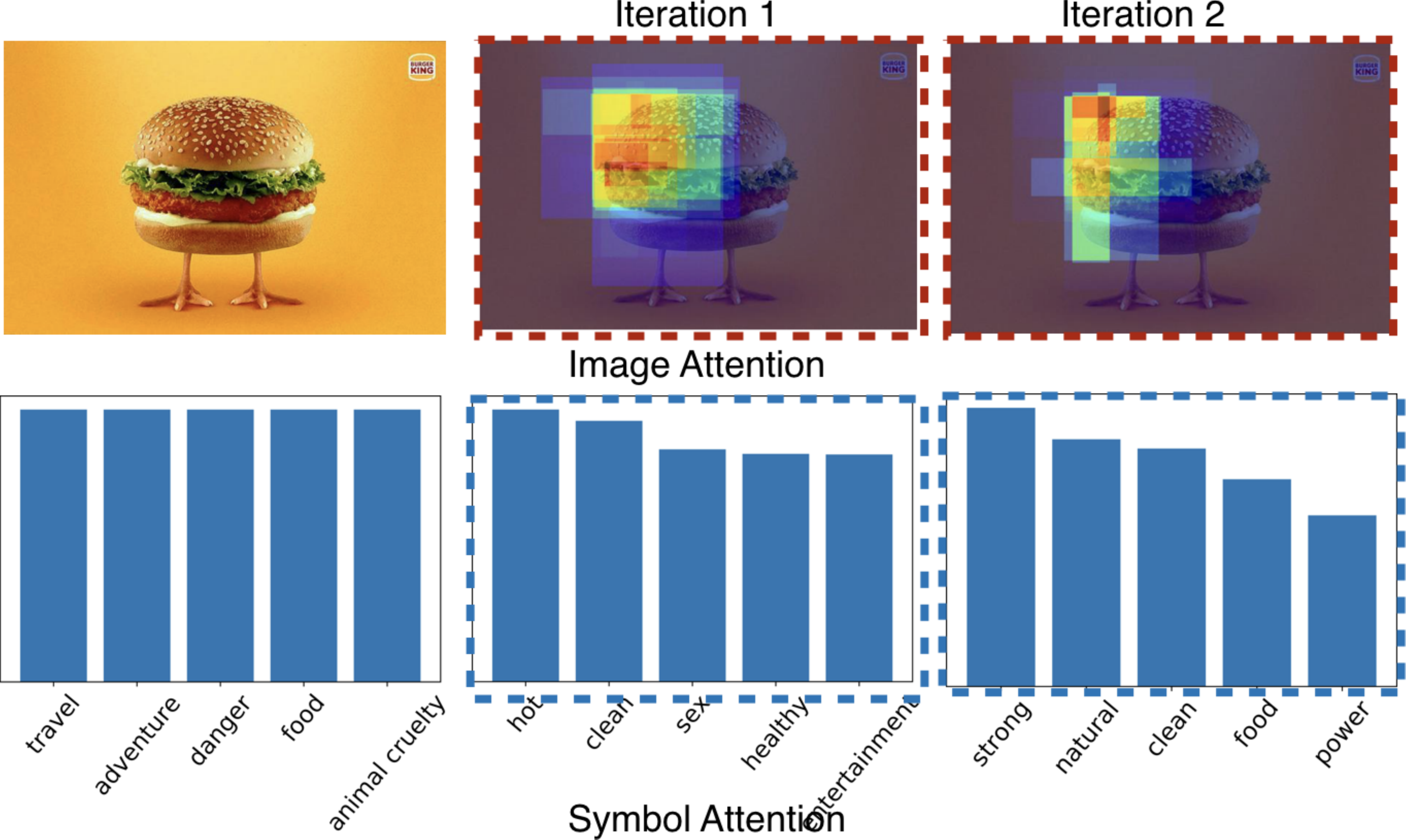} \\
	\includegraphics[width=.5\columnwidth]{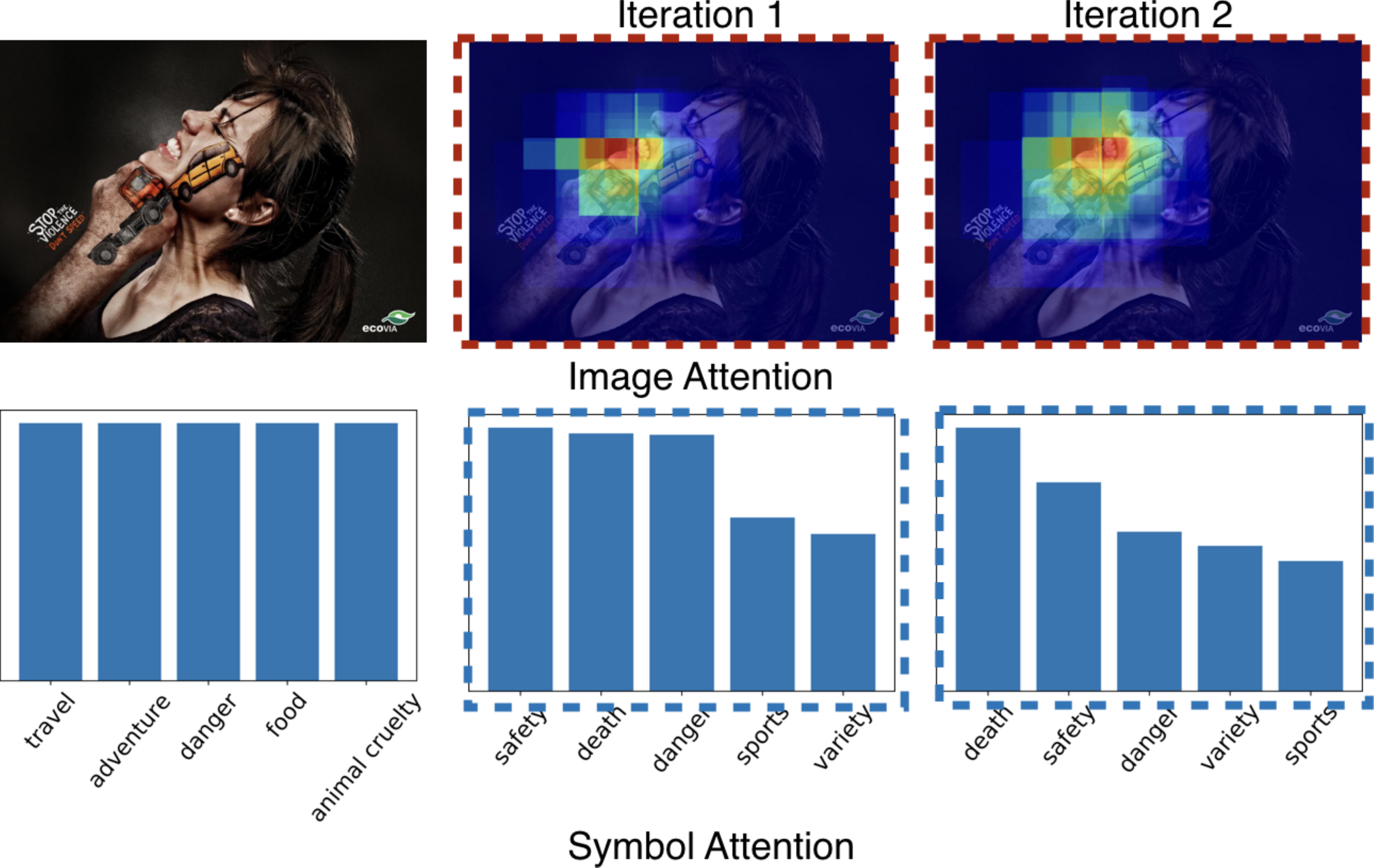} \\
	\includegraphics[width=.5\columnwidth]{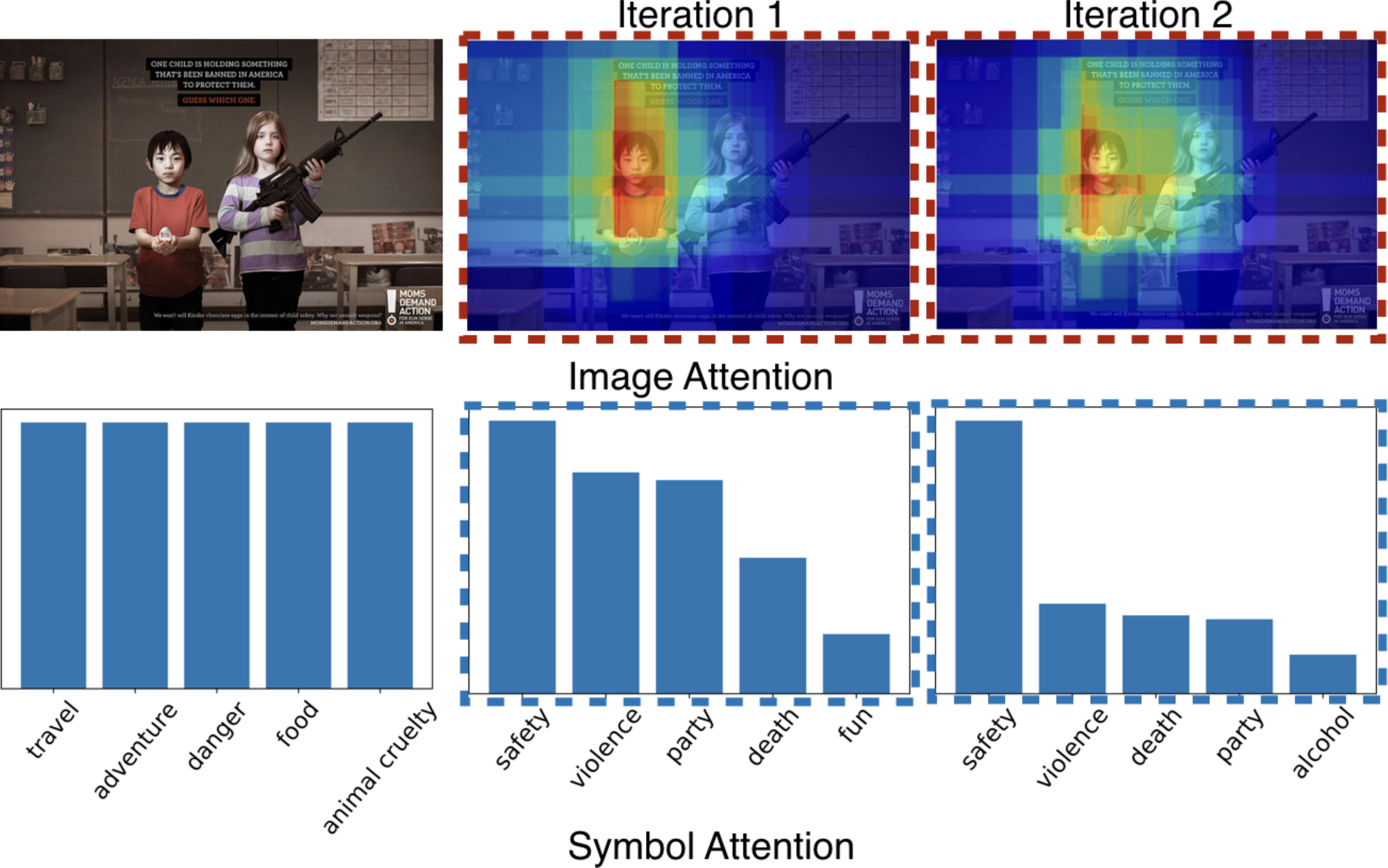} \\
	\includegraphics[width=.5\columnwidth]{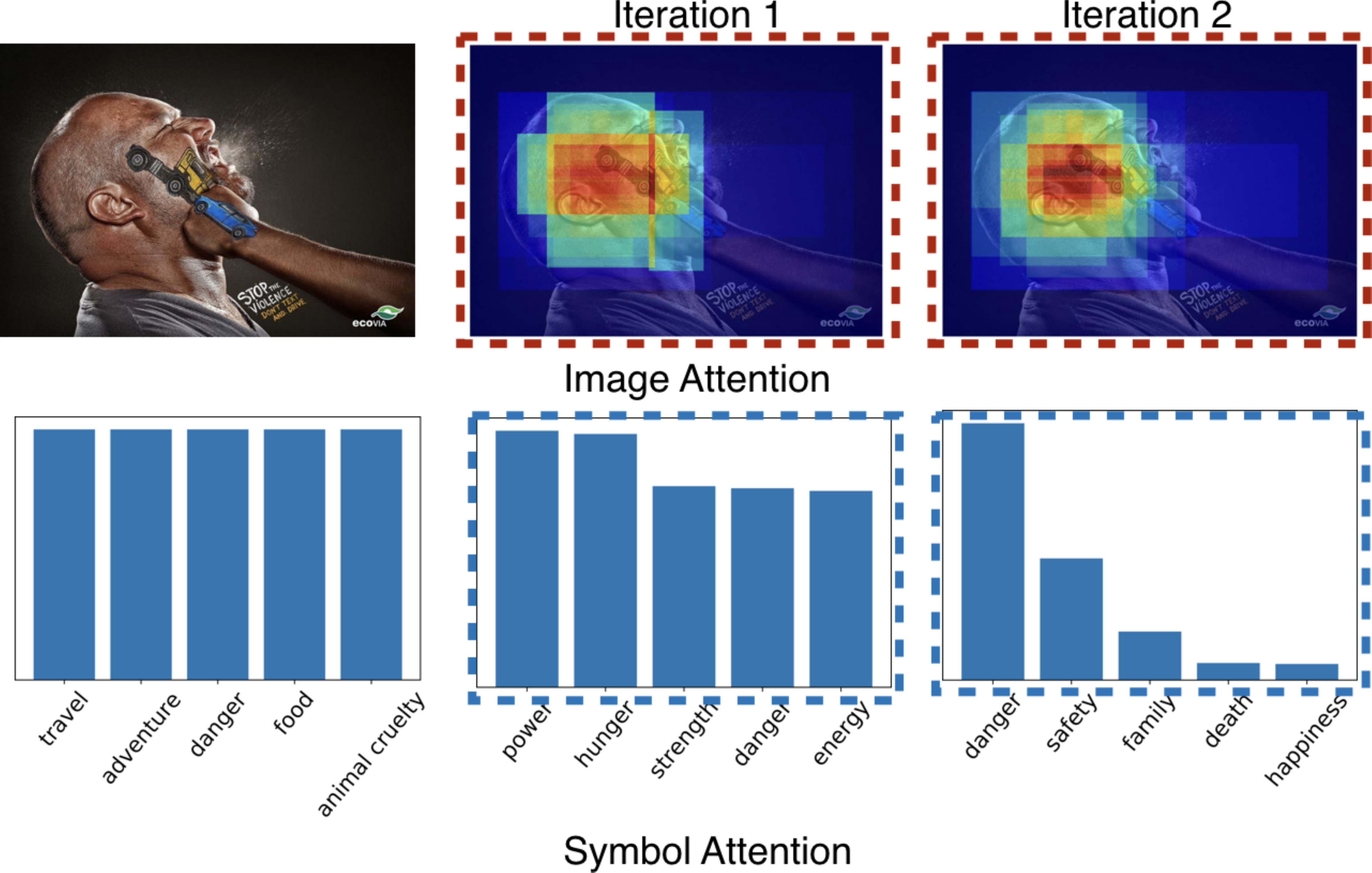} \\
\caption{\small{Figure shows examples of attention scores generated by our algorithm}}
	\label{fig_all}
\end{figure}

\end{document}